%% file: bert_analysis.tex
\documentclass[sigconf]{acmart}

\copyrightyear{2020} 
\acmYear{2020} 
\setcopyright{acmcopyright}\acmConference[SIGIR '20]{Proceedings of the 43rd International ACM SIGIR Conference on Research and Development in Information Retrieval}{July 25--30, 2020}{Virtual Event, China}
\acmBooktitle{Proceedings of the 43rd International ACM SIGIR Conference on Research and Development in Information Retrieval (SIGIR '20), July 25--30, 2020, Virtual Event, China}
\acmPrice{15.00}
\acmDOI{10.1145/3397271.3401195}
\acmISBN{978-1-4503-8016-4/20/07}

\usepackage{booktabs}
\usepackage{color}
\usepackage{balance}
\usepackage{multirow}
\usepackage{colortbl}
\usepackage{arydshln}
\usepackage{multicol}
\usepackage{graphicx}
\usepackage{mathtools}
\usepackage{caption} 
\AtBeginDocument{%
	\providecommand\BibTeX{{%
			\normalfont B\kern-0.5em{\scshape i\kern-0.25em b}\kern-0.8em\TeX}}}

\begin{document}
\fancyhead{}
	
	\title{A Pairwise Probe for Understanding BERT Fine-Tuning on Machine Reading Comprehension}
	
	
    
	\author{Jie Cai}
	\email{caijie@pku.edu.cn}
	\affiliation{%
	  \institution{Peking University}
	}
	
	\author{Zhengzhou Zhu}
	\email{zhuzz@pku.edu.cn}
	\authornote{Corresponding author}
	\affiliation{%
	  \institution{Peking University}
	  }
	
	\author{Ping Nie}
	\email{ping.nie@pku.edu.cn}
	\affiliation{%
	  \institution{Peking University}
	}
	
	\author{Qian Liu}
	\email{Qian.Liu-9@student.uts.edu.au}
	\affiliation{%
	  \institution{University of Technology Sydney}
	}
	
	
	
	
	
	%
	
	
	\begin{abstract}
		 Pre-trained models have brought significant improvements to many NLP tasks and have been extensively analyzed. But little is known about the effect of fine-tuning on specific tasks. Intuitively, people may agree that a pre-trained model already learns semantic representations of words (e.g. synonyms are closer to each other) and fine-tuning further improves its capabilities which require more complicated reasoning (e.g. coreference resolution, entity boundary detection, etc). However, how to verify these arguments analytically and quantitatively is a challenging task and there are few works focus on this topic. In this paper, inspired by the observation that most probing tasks involve identifying matched pairs of phrases (e.g. coreference requires matching an entity and a pronoun), we propose a pairwise probe to understand BERT fine-tuning on the machine reading comprehension (MRC) task. Specifically, we identify five phenomena in MRC. According to pairwise probing tasks, we compare the performance of each layer's hidden representation of pre-trained and fine-tuned BERT. The proposed pairwise probe alleviates the problem of distraction from inaccurate model training and makes a robust and quantitative comparison. Our experimental analysis leads to highly confident conclusions:
		 (1) Fine-tuning has little effect on the fundamental and low-level information and general semantic tasks.
		 (2) For specific abilities required for downstream tasks, fine-tuned BERT is better than pre-trained BERT and such gaps are obvious after the fifth layer.
		 
		 
	\end{abstract}


    
	%
	

	%
	\maketitle
	
	\section{Introduction}
    	
	In recent years large pre-trained models (e.g. BERT \cite{devlin2019bert}, RoBERTa \cite{Liu2019RoBERTaAR}, XLNet \cite{Yang2019XLNetGA}, ALBERT \cite{Lan2019ALBERTAL}) have brought remarkable advancement to many natural language processing tasks, such as question answering \cite{qu2019bert}, machine translation \cite{imamura2019recycling}, natural language inference \cite{yang2019enhancing}, name entity recognition \cite{sun-yang-2019-transfer}, coreference resolution \cite{kantor2019coreference}, etc. 
	
	Since such large neural models are difficult to explain, many works have analyzed how and why they achieved substantial performances. \citet{lin2019open} and \citet{tenney2019learn} analyze pre-trained models by designing some probing tasks that contain specific linguistic information. \citet{jawahar2019does} visualizing the clustering method based on the representation of each layer in pre-trained BERT. \citet{Clark2019WhatDB} visualize the attention of each head in pre-trained BERT in order to observe the relationship between words. However, few works analyze the impact of fine-tuning, but which is critical to helping us understand its mechanism and thus adapt pre-trained models to complex downstream tasks better. Some existing works \cite{Aken2019HowDB,si2019does} in this field make qualitative analysis rely on additional model training in probing tasks, but accurate quantitative analysis with less dependency is rare. 
	
	To fill the gap, we propose a pairwise ranking approach which is inspired by the observation that many probing tasks involve identifying a matched pair of phrases (e.g. coreference requires matching between an entity and a pronoun). Such quantitative comparison provides high confident conclusions in two aspects:
	(1) It does not involve model training for probing task, removing possible distraction caused by inaccurate model training;
	(2) The pairwise metric only concerns with relative ranking comparison, leading to more robustness compared to absolute distance comparison.

    \begin{table}[htbp]
    \caption{Probing tasks for machine reading comprehension with according examples. Underlined spans within the same question-context example are defined as matching pairs.}
 	\label{tb:examples}
    \begin{tabular}{l}
    \hline
    \textbf{Synonyms } \\
    Question: \textit{what kind of \underline{political system} does Spain have?}\\
    Context: \textit{…\underline{The form of government} in Spain is a parliamentary} \\\textit{monarchy, that is, a social…}\\
    Answer: \textit{a parliamentary monarchy} \\
        \hline
    \textbf{Abbreviation}\\
    Question: \textit{what happens to the \underline{rbc} in acute hemolytic reaction?}\\
    \begin{tabular}[c]{@{}l@{}} Context: \textit{…as it results from rapid destruction of the donor \underline{red}}\\ \textit{\underline{blood cells} by host antibodies (IgG , IgM)…}\end{tabular}\\
    Answer: \textit{rapid destruction of the donor red blood cells by host} \\\textit{antibodies} \\
        \hline
    \textbf{Coreference}\\
    Question: \textit{when did locked out of heaven come out?}\\
    \begin{tabular}[c]{@{}l@{}}Context: \textit{…``\underline{Locked Out of Heaven}'' is a song … \underline{It} was released}\\ \textit{as the lead single from the album on October 1, 2012}…\end{tabular} \\
    Answer: \textit{October 1, 2012} \\
        \hline
    \textbf{Answer Type \& Boundary }\\
    Question: \textit{\underline{when} does the movie battle of the sexes come out?}\\
    \begin{tabular}[c]{@{}l@{}} Context: \textit{…released in the United States on \underline{September 22, 2017},}\\ \textit{by Fox Searchlight Picture…}\end{tabular}\\
    Answer: \textit{September 22, 2017} \\
    \hline
    \end{tabular}
    \vspace{-1.5em}
    \end{table}
	
     In this paper, we take the machine reading comprehension (MRC) task as an example and demonstrate that the proposed pairwise ranking approach is a good way to analyze the impact of fine-tuning. The MRC task is defined as follows: Given a question $q$ and a context $c$, a MRC system extracts a span from $c$ as the answer $a$ according to $q$. Intuitively, given a question, the MRC system needs to first locate to relevant sentence and identify the boundary of $a$. To achieve these goals, we identify five phenomena in MRC:

	\begin{itemize}
		\item \textbf{Synonyms identification} is required to locate a relevant sentence. In order to answer the corresponding question in Table \ref{tb:examples}, the model needs to know that ``\textit{the form of government}" in context is a synonym of ``\textit{political system}".
		
		\item \textbf{Abbreviation identification} is another capability required to locate a relevant sentence. For the corresponding example in Table \ref{tb:examples}, in order to answer the question, the model needs to know that ``\textit{rbc}" in the question semantically matches ``\textit{red blood cell}" in context. 
		
		\item \textbf{Coreference resolution} is closely related to the question. We guarantee the pronoun must be relevant to the answer (pronoun appears in the sentence where the answer locate) and the entity referred to by the pronoun appears in both the question and the context. An example is shown in Table \ref{tb:examples}. ``\textit{locked out of heaven}'' is the entity, the pronoun ``\textit{it}" in the sentence that answer locates refers to the entity ``\textit{locked out of heavens}" in the context.
		
		\item \textbf{Answer type} A good MRC system will determine the type of entity based on the words in the question(where, who, when...) to determine the answer and extract it from the related sentence.
		
		\item \textbf{Boundary detection} is necessary for answer extraction. A good MRC system will determine the answer by detecting the boundary and then extract it from the located sentence. We analyze the ability of the model by outputting the logits score of each position.
	\end{itemize}
	
	\begin{table}[]
    \caption{BERT-\small{base} performance on NQ dataset.}
    \label{tb:performance}
    \resizebox{0.95\linewidth}{!}{%
    \begin{tabular}{@{}lllllll@{}}
    \toprule
                         & \multicolumn{3}{c}{\textbf{Long answer}} & \multicolumn{3}{c}{\textbf{Short answer}} \\ 
    Model               & P         & R        & F1       & P         & R         & F1       \\ \midrule
    BERT (pre-train) & 16.81     & 28.10    & 21.03    & 4.16     & 3.81     & 4.57    \\
    \midrule
    BERT (fine-tune) & \textbf{56.91}     & \textbf{62.30}    & \textbf{59.48}    & \textbf{46.74}     & \textbf{40.25}     & \textbf{43.25}    \\ \bottomrule
    \end{tabular}
    }
    \end{table}
    
	In the above five phenomena, \textbf{Abbreviation} and \textbf{Synonyms} are two kinds of general capabilities, not unique to the MRC tasks. In other words, these two capabilities can be learned by some common tasks. As for \textbf{Coreference}, \textbf{Answer type} and \textbf{Boundary}, they are unique to span extraction MRC task which means these three capabilities are essential for doing the span extraction MRC task. 
	
	In order to probe the above phenomena, we design a pairwise ranking metric to quantitatively compare pre-trained and fine-tuned model with in-domain data. 
	The metric is designed to measure whether matching pairs are closer than random un-matching pairs that aim to provide insight about how well related information are encoded\footnote{Our source code is available at https://github.com/hitnq/analyse\_bert.}.
	This paper makes the following contributions:
	\begin{itemize}
	\item We propose a pairwise probe to understand BERT fine-tuning on the MRC task. The probe involves five phenomena in the MRC task and a pairwise ranking metric to quantitatively analyze impact of fine-tuning BERT on the MRC task.
	\item We prove that the information learned by BERT during  fine-tuning is closely related to the MRC task. Some general linguistic abilities that pre-trained model has not been improved after fine-tuning.
	\end{itemize}


	
	
	\begin{figure*}[!htb]
		\centering
		\includegraphics[width=\linewidth]{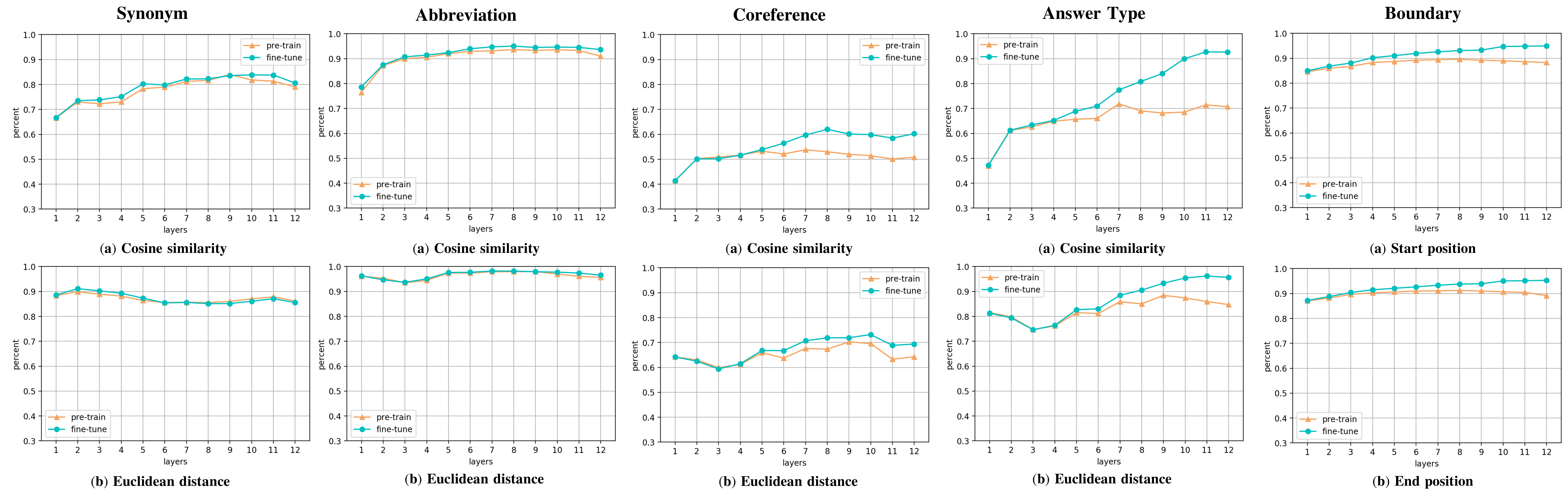}
		\caption{Comparison of pairwise ranking percentage of pre-trained and fine-tuned BERT model on five types of probing tasks. X-axis represents different layers of BERT, y-axix represents pairwise ranking percentage. Top layer lists results with cosine similarity, and bottom layer lists results with Euclidean distance.}
		\vspace{-0.5em}
		\label{fig:result}
	\end{figure*}

	\section{Methods}
	
	We compare pre-trained BERT and the fine-tuned BERT on a large MRC dataset - Natural Questions (NQ) \cite{kwiatkowski2019natural}. The results that shown in Table \ref{tb:performance} indicate fine-tuned BERT outperforms pre-trained BERT by a signiﬁcant margin on the task. In order to investigate what BERT learned during fine-tuning, we conduct comprehensive case studies and summarize five phenomena in the MRC task. We derive according to test data for each skill and leverage a pairwise ranking metric to compare pre-trained and fine-tuned models.

	\subsection{Datasets}
	 The training set of NQ has 307,373 examples, the dev set has 7,830 examples and the test set has 7,842 examples. Each example in NQ contains a question and a corresponding Wikipedia page. All questions in NQ presented are real queries issued to Google's search engine. Answers in each example are annotated with a paragraph (long answer) that answers the corresponding query in the Wikipedia page, while one or more short spans in the paragraph contain the actual answer (short answer). The short answer could be empty. If so, the annotated paragraph is the answer. And if both the long and short answers are empty, then there is no answer on this page. In addition, about 1\% of questions can be answered with ``yes'' or ``no'', so the short answer in the example is marked as ``yes'' or ``no''.

	
    
    
    
	\subsection{Probing Data}
	Our pairwise ranking tasks start by constructing challenge datasets from NQ. Each of our tasks consists of examples that include a matching pair and some un-matching pairs.
	
	\textbf{Synonyms.} A word in query and another word in paragraph (the paragraph that answer is located) are a synonymous pair in WordNet\footnote{https://wordnet.princeton.edu/}. 
	\textbf{Abbreviation.} We ensure that one word in the query is an abbreviation of consecutive phrases in context.
	\textbf{Coreference.} Our goal is to find a coreference pair that related to the question. First, we identify the entity that exists in the question and paragraph (ensure that the entity does not appear in the sentence where the answer is). Then we find the pronoun in the sentence where the answer locate. The entity and pronoun are taken as the matching pair (see Table \ref{tb:examples}).
	\textbf{Answer Type.} We extract examples that question starts with ``who'', ``when'' and ``where'' because these questions has specific answer types.
	\textbf {Boundary.} It contains 10k train data and 5k test data from NQ. We simplify the entire paragraph to the sentence where the answer locate. 
	
    When build un-matching pairs, we keep question words in the matching pair unchanged and treat each word in the paragraph that does not appear in the matching pair as another word in the un-matching pair. Suppose the number of words in a paragraph is $para\_len$ and the number of words in the matching pair of the paragraph is $n$. For a matching pair, we construct $\textit{para\_len-n}$ un-matching pairs. 

	\subsection{Pairwise Ranking Metric}
	
	
	
	
	The model we aim to analyze is BERT-{\small base}\footnote{https://github.com/huggingface/transformers} which consist of typically 12 layers of identically sized Transformer\cite{Vaswani2017AttentionIA} and embedding size $d_E$ is 768. We can get independent pair representations $w_1$ and $w_2$ in each layer ($w_1,w_2 \in \mathbb{R}^{1\times d_E}$). We take the average of these tokens' representations if a word is separated to subtokens by BERT's tokenizer. Then we calculate the similarity (cosine similarity and Euclidean distance) of a matching pair (\textit{pos\_sim}) and compare it with the similarity of un-matching pairs (\textit{neg\_sims}) to show whether the layer has the ability to distinguish matching pair and un-matching pair. For each example, we suppose the number of words in corresponding paragraph is $para\_len$ and $i \in [0,para\_len)$. 
	\begin{equation}
    count[i]=\left\{
    \begin{aligned}
    1,\ & \textbf{if} \ neg\_sims[i] < pos\_sim \\
    0,\ & \textbf{if} \ neg\_sims[i] >= pos\_sim 
    \end{aligned}
    \right.
    \end{equation}
    \begin {equation} 
	example\_count = \textbf{SUM}(count)
	\end {equation}
	
	At last, we calculate the percentage that the similarity of matching pair is higher than un-matching pair in all examples. $n$ represents the number of examples and $j \in [1,n]$.
	\begin {equation} 
	percentage = \frac{\sum_{j=1}^{n}{example\_count[j]}}{\sum_{j=1}^{n}{para\_len[j]}}
	\end {equation}
	
	\section{Results}
	Figure \ref{fig:result} compares pre-trained and fine-tuned model in terms of pairwise ranking percentage. 
	We have the following observations and analysis.
	
	 In both pre-trained and fine-tuned models, almost all representations in high layers show superior capability than lower ones. This phenomenon holds in the cosine similarity evaluation. We think that the different trends in Euclidean distance evaluation are due to differences in evaluation methods.
	 
	 In both methods, the first five layers of pre-trained and fine-tuned model perform almost the same in all probing tasks. This leads us to conclude that fine-tuning has little effect on the information contained in the first five layers and these layers capture more fundamental and low-level information used in fine-tuned models.
	 
	 \textbf{Synonyms \& Abbrevation.} Fine-tuned model performs similarly to pre-trained model in all layers. The synonyms and abbreviation are more basic abilities that are not closely related to the MRC task. This phenomenon illustrates that pre-trained BERT does indeed contain fundamental linguistic knowledge.
	 
	 Therefore we conclude that fine-tuning does not affect basic semantic information. The basic semantic information is reflected in the \textbf{Synonyms} and \textbf{Abbrevation} probing tasks and information contained in the first five layers of BERT.
	 
     \textbf{Coreference \& Answer Type \& Boundary.} These three tasks are all necessary for the span extraction MRC task and only when tasks are done can the MRC task be done well. The fine-tuned model has superior performance than pre-trained model in task-specific capabilities and the gap becomes larger after the fifth layer. It means that what the model learns during fine-tuning is related to downstream tasks and the layers after fifth encode task-specific features that are attributed to the gain of scores.

    \section{Related Work}
	Recently there are many analytical methods that have emerged mainly focus on analyzing pre-trained and fine-tuned models.
	
	\textbf{Pre-trained Model} Visualization is a intuitive analysis method which we can easily find the relationship between words, and we can clearly see the trend of model changes. For example, \citet{jawahar2019does} cluster nouns and phrases based on the representation of each layer in pre-trained BERT and demonstrate that pre-trained BERT captures linguistic and semantic information by a series of probing tasks. \citet{Clark2019WhatDB} visualize the attention of each head in pre-trained BERT to observe the relationship between two words. Another analysis method is to design some probing tasks to check whether each layer of BERT contains relevant knowledge. Probing tasks are usually designed to contain specific information. For example, POS-tagging is a task that contains surface level information and relation classification is a semantic task. A small classification network is added to the embedding of each layer of BERT. This structure is simple so that it won't learn too much knowledge to mainly uses the knowledge that Transformer layer has coded \cite{conneau2018you,Tenney2019BERTRT}. \citet{tenney2019learn} introduced a novel ``edge probing'' method to probe the sub-sentential structure of contextualized word embeddings and applied to nine tasks that encompass a range of syntactic and semantic phenomena.
	
	\textbf{Fine-tuned Model} Some works attempt to analyze what the model learns during fine-tuning. \citet{Aken2019HowDB} adopted some classification and semantic probing tasks, such as NER, relationship classification, etc. They analyzed the performance of different layers after using MRC datasets fine-tune. The authors also used some clustering methods to visualize the closeness between words in specific task. \citet{kovaleva2019revealing} summarize five patterns of attention in BERT after fine-tuned by GLUE tasks and and compare the differences between these patterns in pre-trained and fine-tuned versions. There are also some interesting works in adversarial attacks. It verify the robustness of the model through examples that created with specific interference information. \citet{Rychalska_2018} perturbed questions while holding the context unchanged and inspect coefficients estimated for ground truth class. The authors also explored the importance of each word in a question by gradually deleting words in the question. \citet{Rychalska_2018_verb} verified the importance of verbs in questions by replacing them with their antonyms using WordNet. \citet{si2019does} proposed two novel constructing different attack data method to verify the robustness of the model that fine-tuned by Multiple-Choice Reading Comprehension (MCRC) tasks. \citet{Peters_2019} compared the performance of ``feature extraction'' and ``fine-tuning'' on sentence classification and NLI tasks in GLUE and also compared the effects of different model structures on ``feature extraction'' and ``fine-tuning''.

	\section{Conclusion}
	In this paper, we propose a pairwise ranking metric to quantitatively analyze impact of fine-tuning MRC task on BERT, which is closely coupled with the five phenomena in the task. Equipped with such tool, we have verified that pre-trained BERT already gains fundamental and low-level information. In the \textbf{Synonyms} and \textbf{Abbrevation} probing tasks, there is almost no difference after fine-tuning. Our most surprising finding is that, for \textbf{Coreference}, \textbf{Question Type} and \textbf{Boundary} probing tasks, fine-tuning further improves pre-trained model's capability of conducting task-specific tasks after fifth layer. Moreover, fine-tuning has little effect on the information contained before the fifth layer which verify that fine-tuning does not affect fundamental and low-level information.
	
	In the future, we will apply the pairwise ranking metric to analyze impact of fine-tuning on other tasks. Hopefully such analysis will shed light on more diverse analysis of BERT and thus leading to better understanding of deep learning models.
	
	\section{Acknowledgements}
	This paper was supported by National Key Research and Development Program of China (Grant No. 2017YFB1402400), Ministry of Education "Tiancheng Huizhi" Innovation Promotes Education Fund (Grant No. 2018B01004)

	\input{bert_analysis.bbl}

                \bibliographystyle{ACM-Reference-Format}
	\balance

	
	
	
	
	
	
	
	
	
\end{document}